\documentclass[runningheads]{llncs}
\usepackage[T1]{fontenc}
\usepackage{amsfonts} % 导入所需的数学字体包
\usepackage{graphicx}
\usepackage[table,xcdraw]{xcolor}
\usepackage{hyperref}
\usepackage{graphicx}
\usepackage{color}

\hypersetup{
    colorlinks=true,
    urlcolor=blue,    % URL链接颜色
    citecolor=green   % 引用颜色
}

\begin{document}
\title{Revisiting Cephalometric Landmark Detection from the view of Human Pose Estimation with Lightweight Super-Resolution Head}
\titlerunning{Technical report of the Team SUTD-VLG's challenge method}

\author{Qian Wu\inst{1,3}\and
Si Yong Yeo\inst{2} \and Yufei Chen\inst{3} \and
Jun Liu\inst{1}}

\authorrunning{Wu et al.}
\institute{
  Information Systems Technology and Design (ISTD) Pillar,\\
  Singapore University of Technology and Design (SUTD) \and
  Lee Kong Chian School of Medicine, Nanyang Technological University \and
  College of Electronic and Information Engineering,\\
  Tongji University, Shanghai, China\\
  \email{fivethousand5k@163.com}
}

\maketitle
\begin{abstract}
Accurate localization of cephalometric landmarks holds great importance in the fields of orthodontics and orthognathics due to its potential for automating key point labeling. In the context of landmark detection, particularly in cephalometrics, it has been observed that existing methods often lack standardized pipelines and well-designed bias reduction processes, which significantly impact their performance.
In this paper, we revisit a related task, human pose estimation (HPE), which shares numerous similarities with cephalometric landmark detection (CLD), and emphasize the potential for transferring techniques from the former field to benefit the latter. Motivated by this insight, we have developed a robust and adaptable benchmark based on the well-established HPE codebase known as MMPose. This benchmark can serve as a dependable baseline for achieving exceptional CLD performance.
Furthermore, we introduce an upscaling design within the framework to further enhance performance. This enhancement involves the incorporation of a lightweight and efficient super-resolution module, which generates heatmap predictions on high-resolution features and leads to further performance refinement, benefiting from its ability to reduce quantization bias.
In the MICCAI CLDetection2023 challenge, our method achieves 1st place ranking on three metrics and 3rd place on the remaining one. The code for our method is available at the \href{https://github.com/5k5000/CLdetection2023}{Github}.
\end{abstract}
\section{Introduction}

Cephalometric analysis holds paramount significance within the field of orthodontics~\cite{cureus-surgical,Diagnostics-surgical}. This crucial diagnostic tool involves the meticulous identification of craniofacial landmarks on lateral cephalograms, providing vital insights into a patient's craniofacial condition and significantly influencing treatment planning decisions. However, obtaining reliable landmark annotations on lateral cephalograms presents formidable challenges due to inherent variability in the quality of skull X-ray imaging and the intricate anatomical variations among individuals. Expertise from seasoned medical professionals is often required, and even for experienced orthodontists, manually identifying these landmarks remains a labor-intensive and time-consuming endeavor.

To achieve automatic and precise landmark localization for this task, researchers have proposed various techniques~\cite{wang-tmi,wang-MIA,wang-sci-report,chen2019cephalometric,jiang2022cephalformer}. Wang et al.\cite{wang-MIA,wang-tmi} initially established a well-defined benchmark along with a logical evaluation protocol for the field. Arıkb et al.\cite{arik2017fully} pioneered the application of deep convolutional neural networks for fully automated quantitative cephalometry. Payer et al.\cite{payer2019integrating} introduced a decoupling approach into two simpler tasks, reducing the demand for extensive datasets. Chen et al.\cite{chen2019cephalometric} introduced a Feature Pyramid Fusion module (AFPF) to extract high-resolution and semantically improved features, along with a fusion of heatmap and offsetmap to refine the results. Jiang et al.\cite{jiang2022cephalformer} proposed a transformer-based two-stage framework, demonstrating substantial improvements over previous baselines while maintaining efficiency. In addition to standard settings, Zhu et al.\cite{zhu2021yolo} pioneered the unification of medical landmark detection tasks by training an efficient network on several combined datasets, demonstrating mutual benefits from combining tasks in a single training process. Quan et al.\cite{quan2022images} focused on the few-shot setting, learning with only a few annotated examples, and introduced a method to determine which images to label for greater value. Jin et al.\cite{UDA} proposed an unsupervised domain adaptation framework with self-training, selecting pseudo labels with dynamic thresholds and achieving feature alignment through adversarial learning, reducing the need for extensive annotation while maintaining significant clinical value. Furthermore, Zhu et al.~\cite{zhu2023uod} presented a universal one-shot learning framework with two stages, effectively leveraging domain-specific and domain-invariant features to extract annotated information and establishing a robust benchmark for practical applications.

Despite the significant contributions of the aforementioned methods to the field's development, we emphasize that medical landmark detection, particularly cephalometric landmark detection, still faces several critical issues:
(1) \textbf{Lack of a Universal Design Paradigm}: There is a lack of a universal design paradigm for these methods. Drawing inspiration from the remarkable success of nnUNet~\cite{nnunet} in medical image segmentation, it becomes evident that a universal and high-quality framework not only serves as an effective benchmark for subsequent work~\cite{nnunet-allyourneed,ye2022exploring} but also stimulates innovative thinking in design~\cite{huang2023stunet}. Therefore, it is imperative to introduce a similar benchmark to the community to foster future research.
(2) \textbf{Attach minor importance on quantization bias}: Many methods overlook the importance of details, especially the biases introduced during the quantization process, which includes data preprocessing and postprocessing. These details often have a noticeable impact on performance and the accuracy of results.

In this paper, we begin by reviewing the classic design paradigm of human pose estimation, along with its meticulously crafted unbiased data processing techniques. Subsequently, we delve into why these approaches should be adapted for cephalometric landmark detection. Furthermore, we identify the shortcomings of such a paradigm and elucidate our motivation for enhancing the task. Finally, we present our method for the MICCAI CLDetection2023 challenge, which demonstrates significant performance superiority.

\section{From the perspective of Human Pose Estimation}
Human pose estimation serves the crucial purpose of accurately inferring the precise joint positions of the human body from images or videos, facilitating a deeper comprehension of a person's posture and movements at various time intervals. In recent years, this field has garnered exceptional attention from researchers in the domain of computer vision, as evidenced by notable contributions~\cite{liujiueccv2016spatio,liujun2019TPAMI,LiujunTPAMI2020,qu2022nips,gong2023diffpose}.

The primary objective of human pose estimation training can be formally stated as follows: Given a dataset denoted as $D$, containing labeled examples of human poses, each comprising an input image or frame $I$ and the corresponding ground-truth key points $P_{gt}$, the aim is to train a neural network to learn a function $F$ characterized by trainable parameters $\theta$. This function approximates the mapping from input images $I$ to predicted key points $P_{pred}$, where $P_{pred} = N(I;\theta)$. During training, the objective is to determine the optimal values of $\theta$ that minimize the prediction deviation across the entire dataset $D$:
 $$
 \theta^{*} = argmin_{\theta}\sum_{(I,P_{gt})\in D}L(N(I;\theta),P_{gt})
 $$
Here, $L$ represents the loss function. Subsequently, these skeletal representations find applications in various tasks, such as action recognition~\cite{wang2021dear}. The above formulation underscores the essence of human pose estimation, which revolves around the concept of `multi-person, multi-frame, landmark detection.'

On one hand, a prevalent architectural paradigm employed in this field is the backbone-neck-head framework~\cite{mmpose}, where the backbone handles feature extraction, the neck manages feature aggregation, and the head is dedicated to specific tasks, thereby eliminating the need for a secondary refinement stage. However, it's worth noting that such consensus appears to be less emphasized in research related to cephalometric landmark detection. For instance, architectures like the two-stage CNN-Transformer hybrid~\cite{jiang2022cephalformer} and the vanilla UNet-like encoder-decoder structure~\cite{zhu2021yolo} have been explored in this context.

\begin{figure}[t]
\includegraphics[width=\linewidth]{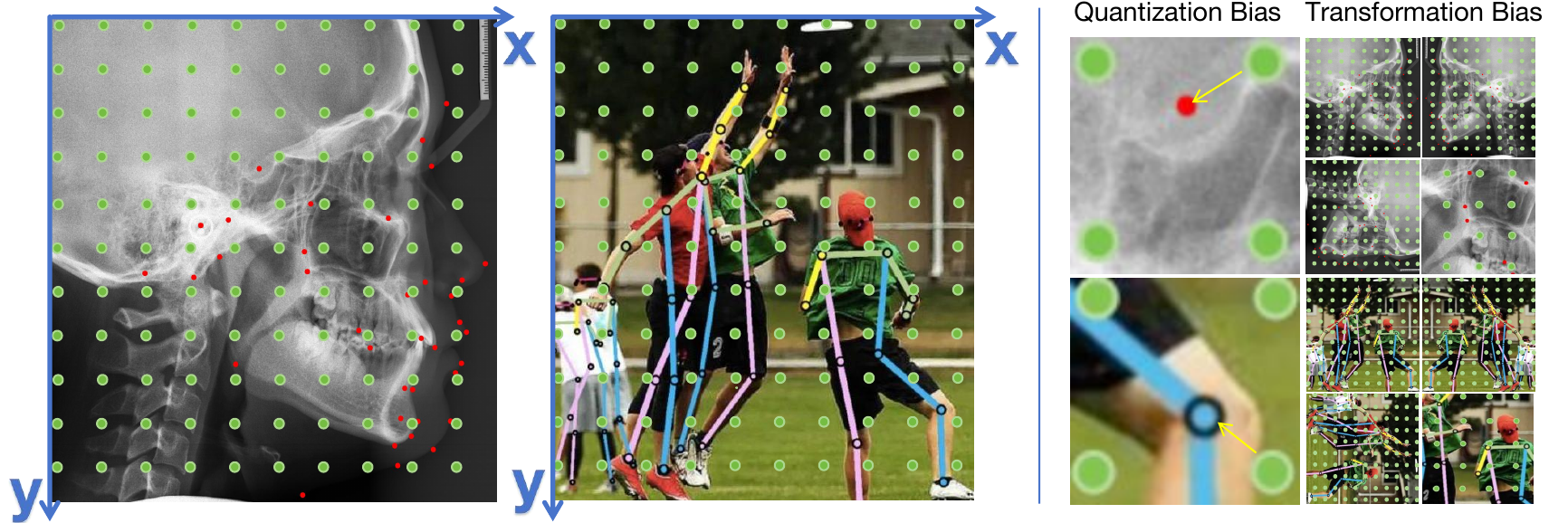}
\caption{In the figure above, we illustrate the challenges related to quantization (left) and bias (right) shared by both cephalometric landmark detection and human pose estimation. The left image showcases quantified points marked with green dots, along with the localization of ground truth denoted by red dots in cephalometric X-ray images and joints in photos. A misalignment could be viewed that most ground truth landmarks cannot fall exactly on the grid points. This misalignment arises because heatmap predictions are often based on downsampled features, as opposed to utilizing features of their original sizes for efficient computation. Therefore, when mapping back to the original resolution, there tends to be always a distance between the nearest predicted grid point to the real localization, as shown in the Quantization Bias column where yellow cursors indicate the gap. Besides, vanilla image transformations like flipping, rotation, and cropping could also lead to quantization bias~\cite{udp}. }

\label{fig2: Quantization and Bias}
\end{figure}

On the other hand, as mentioned earlier, a notable discrepancy exists between human pose estimation and cephalometric landmark detection concerning the importance attributed to `quantization deviation.' In the former field, it has been demonstrated that this factor significantly impacts performance. Pioneering works such as DARK [54] and UDP [55] have elucidated how resolution reduction in networks and data preprocessing operations can introduce bias in quantization. A visual representation of this phenomenon is presented in Figure~\ref{fig2: Quantization and Bias}. To address such issues, DARK explores the design limitations of standard coordinate decoding methods and introduces a more principled distribution-aware decoding approach. Additionally, by generating unbiased heatmaps to enhance the standard coordinate encoding process (i.e., transforming GT coordinates into heatmaps), a novel distribution-aware keypoint coordinate representation method named Distribution-Aware Coordinate Representation of Keypoints (DARK) is proposed by amalgamating these two approaches. The authors of UDP point out that standard data processing primarily involves coordinate system transformations and keypoint format conversions. They observed that conventional flips resulted in inconsistencies compared to the original inference results. Furthermore, statistical errors are present in the keypoint format conversion process. They introduce Unbiased Data Processing (UDP), which encompasses unbiased coordinate system transformations and unbiased keypoint format conversions. Interested readers are recommended to refer to the papers [54, 55] for further details.

Another perspective is that upscaling feature maps and performing heatmap predictions based on higher resolutions can help mitigate biases while preserving valuable high-resolution features for fine-grained prediction and small objects~\cite{zhang2021estimating,srpose}. Nevertheless, the computational load associated with upsampling layers can be substantial (e.g., resource-intensive deconvolution layers), especially when aiming to restore feature maps to their original size to enhance performance.

In summary, building upon established techniques in human pose estimation, our goal is to transfer the following designs into the field of cephalometric landmark detection: (1) unbiased data processing, which ensures that training deviations arise solely from the model itself. (2) A unified backbone-neck-head framework design that incorporates the strengths of previous designs and can serve as a robust foundational model for cephalometric landmark detection. (3) Predicting heatmaps based on higher-resolution feature maps with computationally efficient design.

\section{Method}

\begin{figure}[t]
\includegraphics[width=\linewidth]{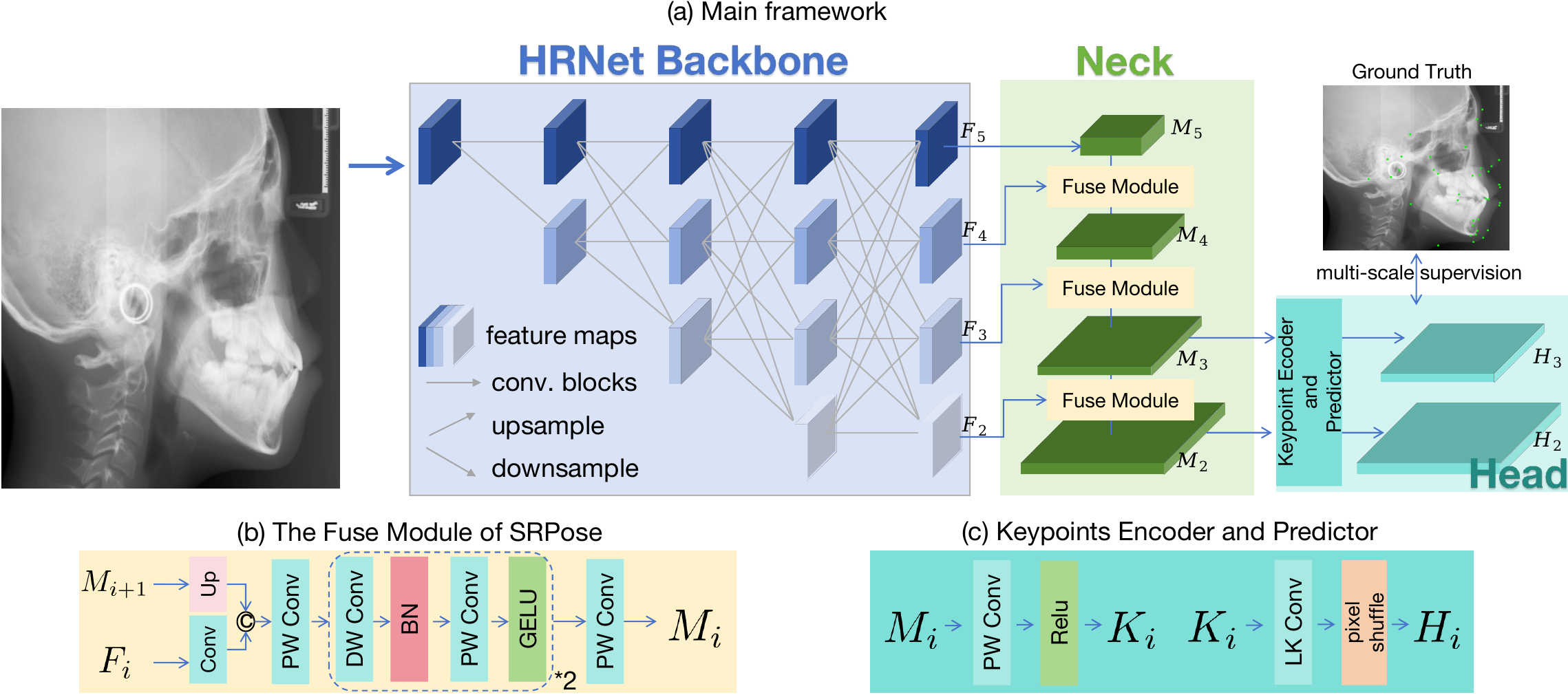}
\caption{The primary framework of our approach for the MICCAI CLdetection23 challenge, along with the details of the key components. (a) The framework adheres to the Backbone-Neck-Head design paradigm. Specifically, we employ the HRNet architecture as the backbone and adapt the design principles from SRPose~\cite{srpose} for the neck and head. (b) The Fuse module aggregates and merges features from different levels, employing efficient operations such as Point-Wise Convolution (PW Conv) and Depth-wise Convolution (DW Conv). (c) The Keypoints Encoder and Predictor receive the fused feature from the neck and project it into heatmaps with super-resolution. Subsequently, predicted multi-scale heatmaps are compared with ground truth landmarks for supervision.}
\label{fig2:main_framework}
\end{figure}

\subsection{Network Architecture}
The overarching framework of our method is illustrated in Figure~\ref{fig2:main_framework}.
\subsubsection{Backbone}
We leverage the widely adopted HRNet~\cite{hrnet} as our backbone to effectively extract features at multiple scales. Given an input image $I \in \mathbb{R}^{H\times W\times3}$, the backbone extracts features as follows:
$$F_2, F_3, F_4, F_5 = B(I)$$
Here, $B$ represents the backbone, and $F_i$ denotes the feature map with a resolution of $\frac{H}{2^i}\times\frac{W}{2^i}$. These features are later fused and aggregated in the neck.

\subsubsection{Neck}
To efficiently fuse features from multiple levels, we adopt the design principles of SRPose~\cite{srpose}, which optimizes model parameters using a series of separable convolutions~\cite{xception,mobilenets}. As depicted in Figure~\ref{fig2:main_framework}(b), the Fuse module initially concatenates $F_i$ and $M_{i+1}$ after passing them through a convolution block and an upsampling operation, respectively. Subsequently, the concatenated features undergo a Point Wise Convolution (PW Conv) block. Following these, two consecutive modules comprising depth-wise convolution, Batch Normalization (BN), PW Conv, GELU activation are applied. Finally, a PW Conv operation yields the fused result. The fusion for each level can be mathematically expressed as:
$$M_i = Fusion(F_i, M_{i+1})$$

\subsubsection{Head}
To mitigate quantization deviations by bridging the resolution gap between the input image and the predicted heatmaps, we follow the design principles of SRPose~\cite{srpose} and Super Resolution techniques~\cite{real}. We encode and upscale the fused features into heatmaps using efficient convolutions and parameter-free pixel shuffle. Specifically, we initially encode the fused feature $M$ for each keypoint as follows:
$$K^0, K^1, K^2, \ldots, K^{N-1} = Encoder(M)$$
Here, $M$ is the fused feature map from the neck, $N$ is the total number of keypoints, and $Encoder$ is a Pixel-wise convolution followed by a RELU function as depicted in Figure~\ref{fig2:main_framework}. For each keypoint, a set of $s^2$ low-resolution heatmaps are generated using a large kernel convolution $f_{LKC}: \mathbb{R}^{1\times h \times w} \rightarrow \mathbb{R}^{(s\times s) \times h \times w}$ as follows:
$$h^{j} = f_{LKC}^{j}(K^j)$$
Here, $f_{LKC}^{j}$ represents the $j$-th large kernel convolution designed for the $j$-th keypoint. These $h^j$ maps are considered as a stack of low-resolution heatmaps for the $j$-th keypoint. Subsequently, $h$ is upsampled using the Pixel Shuffle operation to obtain $H$, a heatmap with the same resolution as the input image $I$:
$$H^{j} = PS(h^j,s)$$
Here, $H^j$ represents the high-resolution heatmap for the $j$-th keypoint, and $PS$ denotes the Pixel Shuffle operation. The parameter $s$ corresponds to the number of heatmaps generated, also serving as the upscaling ratio for the Pixel Shuffle operation.

In this context, we further clarify the heatmap prediction head. As shown in Figure~\ref{fig2:main_framework}, only $M_2$ and $M_3$ are projected and subsequently upsampled to high-resolution heatmaps. We employ this design due to empirical findings that supervision based on tiny-resolution heatmaps, such as $M4$ and $M5$, which are downsampled by a factor of 16 and 32, can sometimes lead to performance degradation. Specifically, $H_3$ and $H_2$ have resolutions of $\frac{H}{4}\times\frac{W}{4}$ and $\frac{H}{8}\times\frac{W}{8}$, respectively, which implies upscaling ratios of 2 and 4 are applied during the Pixel Shuffle operation for $M_3$ and $M_2$.

\subsection{Training Objective}

The training objective is designed to ensure that the predicted multi-scale heatmaps align closely with the corresponding ground truth landmarks. We employ a well-defined loss function to quantify the dissimilarity between the predicted heatmaps and the ground truth. Our objective can be formulated as follows:

\[
\mathcal{L} = \frac{1}{2N} \sum_{j=0}^{N-1} \sum_{i\in\{2,3\}}||H^j_i - G^j_i||_{2}
\]
where $\mathcal{L}$ is the overall loss, $H^j_i$ represents the predicted heatmap at scale $i$ for the $j$-th keypoint, $G^j_i$ denotes the corresponding ground truth gaussian heatmap at scale $i$ for the $j$-th keypoint, $N$ is the total number of keypoints, and $||\cdot||_{2}$ denotes the L2 norm between the predicted and ground truth heatmaps.

\section{Experiments}
\subsection{Datasets}
\begin{figure}[b]
\includegraphics[width=\linewidth]{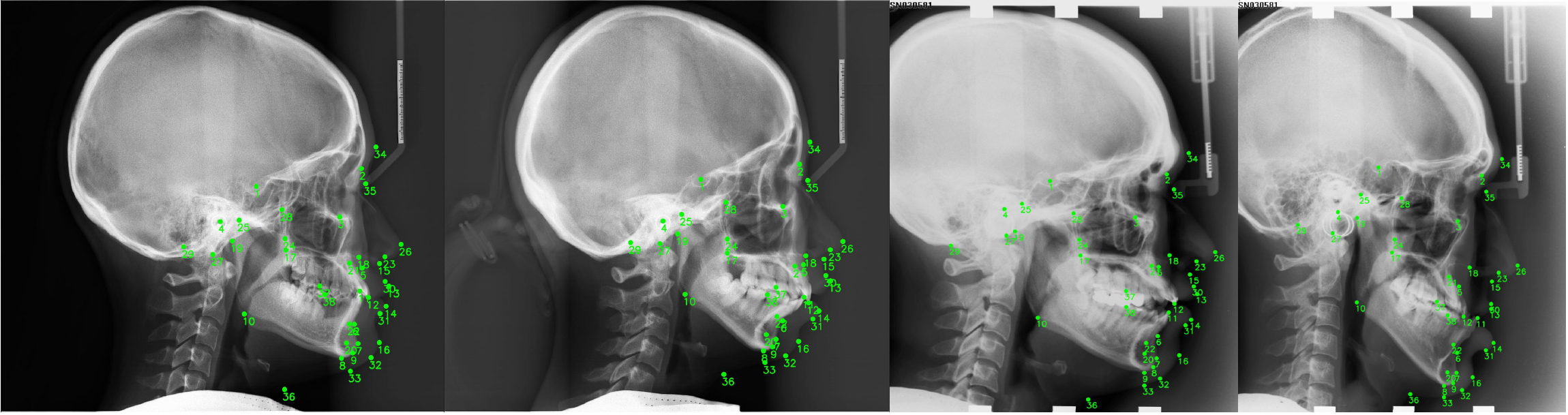}
\caption{A glimpse on the dataset (Best viewed upon zooming in). The green points together its indices represent 38 landmarks annotated for each case.}
\label{fig3: dataset_demo}
\end{figure}

In the context of the MICCAI CLdetection2023 challenge, participants are granted access to an unprecedentedly diverse dataset, which is an expansion built upon existing benchmark datasets. The dataset comprises a total of 600 X-ray images sourced from three distinct medical centers, each meticulously annotated with 38 craniofacial landmarks for every case. A glimpse of the dataset can be viewed in Fig.~\ref{fig3: dataset_demo}. The dataset is officially split into training, validation, and test datasets, with 400, 50, and 150 cases respectively. As participants who could only access the training dataset, we locally split it into training and validation datasets, with 350 and 50 cases respectively.

\subsection{Evaluation Metrics}
The evaluation metrics are the mean radial error (MRE), and the successful detection rate (SDR) under 2 mm, 2.5 mm, 3mm, and 4 mm, following~\cite{wang-tmi,wang-MIA}.
\subsection{Implementation Details}
Our codebase is built upon the widely recognized human pose estimation framework MMPose~\cite{mmpose}. Throughout all experiments and challenge submissions, we have employed HRNet-W48~\cite{hrnet} as the backbone. A value of 6 is set for $\sigma$ to generate Gaussian heatmaps. During the training phase, we utilize data augmentation operations such as RandomFlip, RandomHalfBody, RandomBBoxTransform, and Resize. We employ the Adam optimizer with an initial learning rate of 0.0005 and a total training duration of 100 epochs with a batch size of 2. We incorporate DARK~\cite{dark} as our debiasing strategy.

During the testing phase, we perform heatmap flipping. All experiments were conducted on a machine equipped with 2 RTX 3090 24G GPUs.

\subsection{Quantitative Results}

\subsubsection{Online Results.}
In this part, we report the performance on the official website, with 2 leaderboards and 4 metrics in total: validation leaderboard\footnote{https://cl-detection2023.grand-challenge.org/evaluation/challenge/leaderboard/} and test leaderboard\footnote{https://cl-detection2023.grand-challenge.org/evaluation/testing/leaderboard/}. The results are shown in Table~\ref{tab:online results}. It can be observed that our method achieves top-performing outcomes that we rank 1st on 3 metrics and 3rd on the remaining one.

\begin{table}[htp]
\centering
\caption{The online results of our method. Our results are marked blue. Note that results from the same team are ignored, while only the best ones are kept. Note that, we assemble 7 models for the final submission. }
\begin{tabular}{|cc|cc|}
\hline
\multicolumn{2}{|c|}{Validation}         & \multicolumn{2}{c|}{Test}                                     \\ \hline
\multicolumn{1}{|c|}{SDR2$\uparrow$}    & MRE$\downarrow$ (mm) & \multicolumn{1}{c|}{SDR2$\uparrow$}    & MRE$\downarrow$ (mm)                       \\ \hline
\multicolumn{1}{|c|}{\cellcolor[HTML]{96FFFB}\textbf{77.5263}} &
  \cellcolor[HTML]{96FFFB}\textbf{1.4843} &
  \multicolumn{1}{c|}{\cellcolor[HTML]{96FFFB}\textbf{75.7193}} &
  \textbf{1.5176} \\
\multicolumn{1}{|c|}{77.4737} & 1.4854   & \multicolumn{1}{c|}{75.7193} & 1.5510                         \\
\multicolumn{1}{|c|}{77.2105} & 1.5018   & \multicolumn{1}{c|}{75.6842} & \cellcolor[HTML]{96FFFB}1.5557 \\
\multicolumn{1}{|c|}{76.3684} & 1.5106   & \multicolumn{1}{c|}{75.5263} & 1.5843                         \\
\multicolumn{1}{|c|}{76.0526} & 1.5112   & \multicolumn{1}{c|}{75.2105} & 1.6159                         \\ \hline
\end{tabular}%
\label{tab:online results}
\end{table}
\subsubsection{Offline Results.}
% \usepackage{graphicx}
\begin{table}[htp]
\centering
\caption{The road map of our model selection and hyperparameter tuning process. We are here to clarify how to interpret the table. To begin with, this table should be read from top to bottom. Our exploration is based on B1. Then we progressively add modules or alter the hyperparameters. Once we find a better setting,  we will replace the old benchmark with a new one. For example, when we apply a larger input size $1024\times1024$ and achieve a better score, we then set it as a new benchmark, denoted as 'B2: B1+ larger input ($1024\times1024$)'. All the rows follow such regular. When we notice an alternation could not contribute to the performance, we will leave the old benchmark unchanged. Note that all the new benchmarks are marked grey, and start with B\{n\}. In the end, B10 is the final setting, we used for challenge submission. Here, sigma represents the $\sigma$ used for generating Gaussian heatmaps. OHKM represents Online Hard Keypoints Mining. DA stands for Data Augmentation. Note that the \textbf{effectiveness of the super-resolution design} is shown by comparing the B8,B9 and B10. }
\resizebox{\textwidth}{!}{%
\begin{tabular}{c|ccccc}
\hline
Method                              & MRE$\downarrow$ (mm)    & SDR2$\uparrow$   & SDR2.5$\uparrow$  & SDR3$\uparrow$   & SDR4$\uparrow$    \\ \hline
\rowcolor[HTML]{EFEFEF} 
B1: HRNet-w48                       & 2.61   & 65.211 & 75.842 & 82.158 & 90.1054 \\
+ adaptive loss~\cite{wang2019adaptive}                     & 2.358  & 64.895 & 75.368 & 82.947 & 90.737  \\
\rowcolor[HTML]{EFEFEF} 
B2: B1+ larger input ($1024\times1024$)    & 2.496  & 70.158 & 78.579 & 84.211 & 90.947  \\
\rowcolor[HTML]{EFEFEF} 
B3: B2 + sigma=3                    & 1.893  & 71.895 & 79.842 & 85.105 & 91.526  \\
\rowcolor[HTML]{EFEFEF} 
B4: B3 +more DA                     & 1.796  & 72.737 & 80.526 & 85.316 & 92      \\
\rowcolor[HTML]{EFEFEF} 
B5: B4 + longer (300 epoch)                     & 1.786  & 72.895 & 80.263 & 86.105 & 92.579  \\
\rowcolor[HTML]{EFEFEF} 
B6: B5 + DARK~\cite{dark}                       & 1.729  & 74.158 & 81.368 & 86.474 & 92.632  \\
+ Focal Loss~\cite{focalloss}                       & 1.827  & 68.895 & 78.789 & 85.895 & 93.105  \\
+ OHKM~\cite{onlinehard}                              & 1.924  & 69.684 & 78.316 & 85.842 & 92.263  \\
+ with larger heatmap($1024\times1024$)         & 2.96   & 74.368 & 79.895 & 84.684 & 89.842  \\
sigma = 8                           & 1.763  & 75     & 81.632 & 86.684 & 92.737  \\
\rowcolor[HTML]{EFEFEF} 
B7: B6 + flip test                  & 1.783  & 76     & 82.947 & 87.632 & 92.789  \\
sigma = 10                          & 1.815  & 73.421 & 80.526 & 86.053 & 92.632  \\
\rowcolor[HTML]{EFEFEF} 
B8: B7 + sigma = 6                  & 1. 702 & 76.737 & 83.789 & 88.632 & 93.321  \\
sigma=4                             & 1.731  & 76.895 & 83.263 & 87.211 & 92.316  \\
input size (800,640)                & 1.65   & 74.684 & 83.216 & 87.105 & 92.789  \\
Sigmoid before MSE Loss             & 50.24  & 3.368  & 3.895  & 4.105  & 4.684   \\
\rowcolor[HTML]{EFEFEF} 
B9: B8+SRpose~\cite{srpose} (supervise 4 scales)  & 1.65   & 76.524 & 82.539 & 87.928 & 92.342  \\
SRPose (supervise 3 scales)         & 1.61   & 76.947 & 82.947 & 88.053 & 92.684  \\
\rowcolor[HTML]{EFEFEF} 
B10: B9+SRPose (supervise 2 scales) & 1.589  & 77.632 & 83.684 & 87.737 & 92.579  \\
SRpose (supervise 1 scale)          & 1.831  & 77.474 & 83.053 & 88.348 & 92.684  \\ \hline
\end{tabular}%
}
\label{tab:offline results}
\end{table}
As there is a submission limit for participants, we have to carry out a series of experiments locally to determine the best settings. We present a road map of our hyperparameter tuning and model selection in Table~\ref{tab:offline results}.

\subsection{Qualitative Results}
\begin{figure}[!htp]
\includegraphics[width=\linewidth]{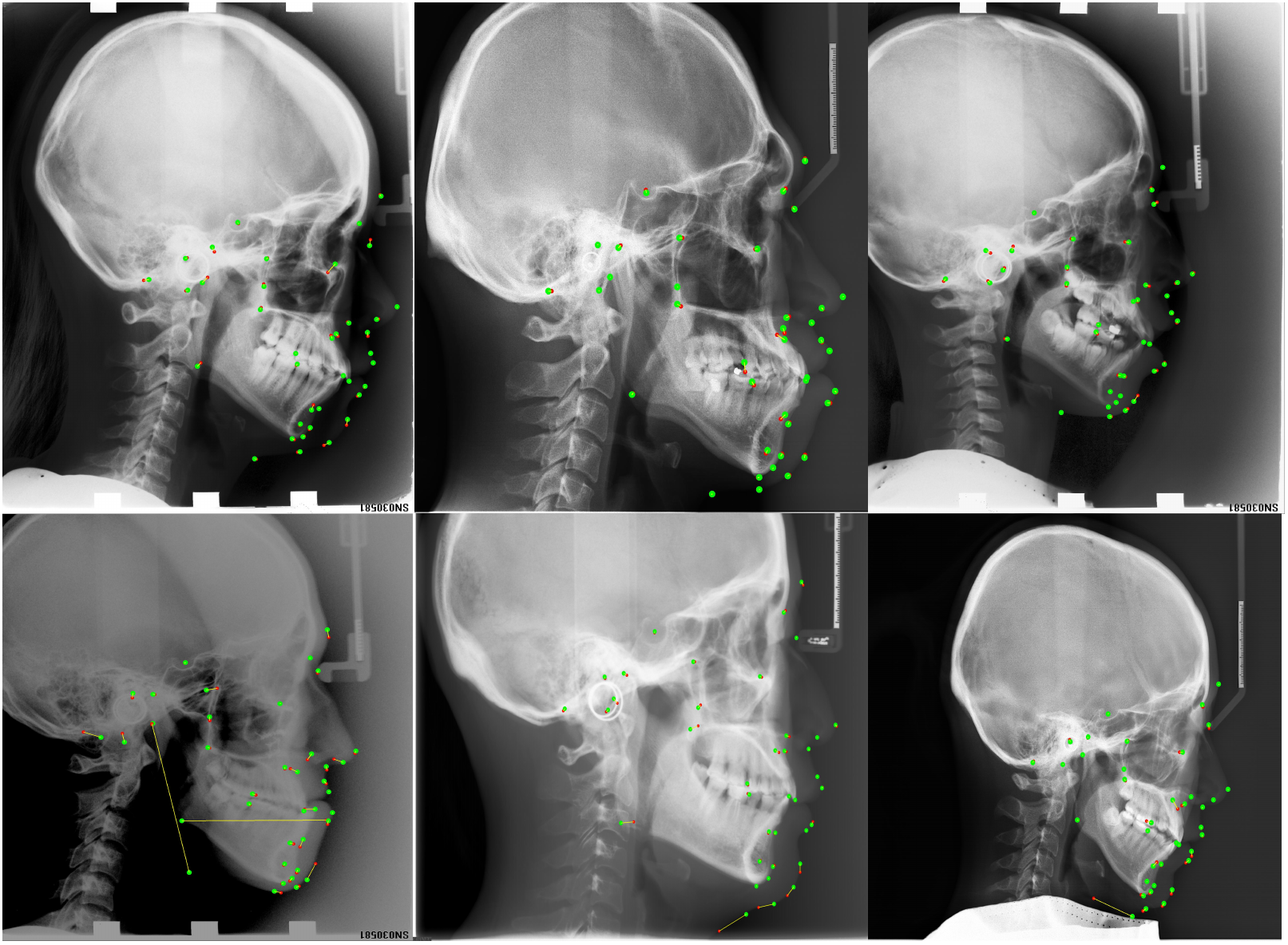}
\caption{Some qualitative results of our method. The predicted landmarks and ground truth are marked in red points and green points, respectively. The distance of the prediction and corresponding ground truth are marked with yellow lines. Note that we present good cases in the first row and bad cases in the second row.}
\label{fig4: visualization}
\end{figure}

The visualization of some cases of our method is shown in Fig.~\ref{fig4: visualization}

\section{Discussion}
From the results presented in Table~\ref{tab:offline results}, it is evident that the performance of our model is sensitive to crucial parameters, including input image size and the $\sigma$ value for Gaussian maps. By carefully tuning these fundamental hyperparameters and modules, we achieved a noteworthy improvement of approximately 10 in SDR2. This underscores the significance of a meticulous parameter-tuning process. It is noteworthy that SRPose~\cite{srpose} also contributes to enhancing performance when configured with the appropriate supervision settings.

In our offline experiments, we conducted an ablation study to investigate the impact of various supervision levels in SRPose. The results of this study, presented in the last four rows of Table~\ref{tab:offline results}, motivated us to retain only the last two supervision levels.

In addition to the visualizations shown in Fig.~\ref{fig4: visualization}, we also analyzed cases that did not perform well in unshown data. We identified several reasons for the failure of certain landmarks: (1) Obscured Edges: Landmarks located around the jaw region often suffer due to obscured edges. (2)Out-of-Distribution Imaging: Some images deviated significantly from the majority in terms of quality or viewing angle. These out-of-distribution cases posed challenges to the model's performance. (3) Disruptive Elements: Objects like clothing could occasionally interfere with the model's ability to accurately predict landmarks.
To mitigate these challenges, we suggest potential directions for improvement: (1) Robust Feature Enhancement: Develop models with features that are more robust and generalizable, capable of handling anomalies and challenging cases.
(2) Incorporate Landmark Reference Standards: Embed models with inherent landmark reference standards, enabling them to perform well in a broader range of scenarios. These directions will be explored in future work, with the aim of further advancing the performance and robustness of our model.
\section{Conclusion}
In this paper, we present our method for the MICCAI CLdetecion2023 challenge. We first revisit the gap between cephalometric landmark detection and human pose estimation, and stress that valuable techniques, which focus on the problem of quantization bias, and the unified designing paradigm could be adapted to cephalometric landmark detection. To further alleviate the deviation issue, we are motivated to adopt a super-resolution design, which performs heatmap prediction based on high-resolution feature maps and is computationally efficient. Then we present our method in detail and present a comprehensive analysis of it. Our methods achieve top-performing performance on the official leaderboards, which ranks 1st on 3 metrics and 3rd on the remaining one.
\section*{Acknowledgement}
We would like to thank the MICCAI CLdetection2023 organizers, for providing well-established baselines and their altruistic service for the contest. We appreciate all the contributors of the MMPose Package~\cite{mmpose}.

%
%
%

\bibliographystyle{splncs04}
\clearpage
\bibliography{reference}
\end{document}